\documentclass[sigconf]{aamas} 

\usepackage{balance} 
\usepackage{moreverb}
\usepackage{float}
\usepackage{algorithm}
\usepackage{algpseudocode}
\usepackage{graphicx}
\usepackage{caption}
\usepackage{subcaption}

\floatstyle{ruled}
\newfloat{pddl}{thp}{lop}
\floatname{pddl}{PDDL listing}

\setcopyright{none}

\acmConference[None]{Technical Report. This is NOT the official paper accepted in AAMAS 2021. Please, cite the AAMAS 2021 paper to reference this work.}
\copyrightyear{}
\acmDOI{}
\acmPrice{}
\acmISBN{}

\title[Optimized Execution of PDDL Plans using Behavior Trees ]{Optimized Execution of PDDL Plans using Behavior Trees }

\author{Francisco Martín}
\affiliation{
  \department{Intelligent Robotics Lab}
  \institution{Rey Juan Carlos University}
  \city{Fuenlabrada, Madrid}}
\email{francisco.rico@urjc.es}

\author{Matteo Morelli}
\affiliation{
  \institution{CEA list}
  \city{Palaiseau, France}}
\email{Matteo.MORELLI@cea.fr }

\author{Huascar Espinoza}
\affiliation{
  \institution{CEA list}
  \city{Palaiseau, France}}
\email{Huascar.ESPINOZA@cea.fr}

\author{Francisco J. R. Lera}
\affiliation{
  \institution{University of Leon}
  \city{Leon, Spain}}
\email{fjrodl@unileon.es}

\author{Vicente Matellán}
\affiliation{
  \institution{University of Leon}
  \city{Leon, Spain}}
\email{vicente.matellan@unileon.es}

\begin{abstract}

Robots need task planning to sequence and execute actions toward achieving their goals. On the other hand, Behavior Trees provide a mathematical model for specifying plan execution in an intrinsically composable, reactive, and robust way. PDDL (Planning Domain Definition Language) has become the standard description language for most planners. In this paper, we present a novel algorithm to systematically create behavior trees from PDDL plans to execute them. This approach uses the execution graph of the plan to generate a behavior tree. The most remarkable contribution of this approach is the algorithm to build a Behavior Tree that optimizes its execution by paralyzing actions, applicable to any plan, taking into account the actions' causal relationships. We demonstrate the improvement in the execution of plans in mobile robots using the ROS2 Planning System framework.

\end{abstract}


\newcommand{\BibTeX}{\rm B\kern-.05em{\sc i\kern-.025em b}\kern-.08em\TeX}

\begin{document}


\pagestyle{fancy}
\fancyhead{}


\maketitle 


\section{Introduction}

In this work we want to improve the execution  of plans and tasks in mobile service robots. Current automated planning and scheduling (Artificial Intelligence (AI) Planning) systems cannot be applied directly to real robots. The AI planning community is a prolific research area, but it was often seen as focused on solving highly abstract problems instead of applying its theory to real-world situations. Classical planning algorithms assume a \emph{closed world}, in which the state of the world changes only in response to the dispatching of individual actions. This assumption does not hold for a service robot deployed in a real environment because the world state changes independently of the plan execution. The world must be continuously perceived and taken into account both in producing plans and in their execution.

Another aspect to consider is how to store the knowledge used in the generation of plans. This knowledge must be accessible and updatable by the plan solver and other elements of the planning system. The execution of the plans is another crucial aspect for AI planning in robotics. Once the plan is generated in a computer, it must be carried out in the real world. This execution includes handing over the actions to the components that must perform them and keep checking the action requirements during their execution. The actions in a plan contain the timestamp in which they must be executed, but this is usually not realistic in the real world, in which the time dedicated to an action can vary.

Existing frameworks for developing robotics software usually include AI plan solvers, such as POPF \cite{popf}, or Fast Downward \cite{fastdownward}. They also include tools for managing the knowledge base, which captures both the domain and the problem to be solved, as well a plan executor. In particular, we will focus on the ROS (Robotic Operating System) framework, which is becoming the {\em de facto} standard for robotic software development.

Currently, the most popular planning framework in ROS is ROSPlan \cite{ROSPlan19}, which encapsulates both planning and dispatching. Unfortunately, this system is only available in ROS 1, which is starting to be deprecated and presents many performance shortcomings and security vulnerabilities. The work presented in this paper has been  developed for the ROS2 Planning System (PlanSys2 in short) that aspires to be the reference framework for AI Planning in the next version of ROS: ROS2. PlanSys2\footnote{\url{https://github.com/IntelligentRoboticsLabs/ros2_planning_system/}} provides a set of services for evaluating different planning algorithms. It also aims to provide management and execution mechanisms that are not present in ROS 1 frameworks, such as multi-robot support, and parallel execution of the planned actions.

In particular, the work described in this paper is focused on the parallel execution of the actions that take part of a plan in PlanSys2. This proposal uses Behavior Trees to execute plans. Behavior Trees have recently gained a lot of traction in the Robotics community \cite{Colledanchise} due to their flexibility in  creating complex behaviors composed of simpler ones. The main assumption of this work is that  Behavior Trees can adequately represent plans, including the causal constraints between the actions, and the evaluation of preconditions. If a plan can be coded as a Behavior Tree, its execution is much easier, without neglecting all the aspects to take into account, like checking for requirements or applying effects with the correct timing. We can automatically include nodes with this purpose in the tree.

The main contribution of this paper is a novel algorithm to build a Behavior Tree that optimizes the execution time of plans by explicitly producing parallel actions. Converting a plan into a behavior tree with a set of purely sequential actions is feasible but sub-optimal. Within a plan, we can identify several execution flows that converge at one point and diverge at another, which can be executed in parallel. Manually creating these behavior trees is tricky, but can be done easily by an expert. Systematic and general algorithms are required to automatically do this work. The algorithm proposed in this paper meets these requirements through two mechanisms. The first one is a new type of node whose function is to wait for the execution of an action running in parallel, or has not yet been executed, on any other side of a tree. The second mechanism is to convert the nodes that execute the action into singletons \cite{Gamma94} to execute the same action from various points in the tree. These two mechanisms greatly simplify the automatic creation of Behavior Trees for optimal execution.

This paper is structured as follows. In Section \ref{sec:related}, the current approaches related to the targeted problem are reviewed. Section \ref{sec:graph} presents some concepts of Planning Definition Language (PDDL) and how plans can be represented as a planning graph. Section \ref{sec:bt} describes our algorithm to build a Behavior Tree from this planning graph. We evaluate our approach in Section \ref{sec:evaluation}, and finally, some conclusions are discussed in Section \ref{sec:conclusions}.

\section{Related Work}
\label{sec:related}

Automatic generation of plans  has been used in Robotics for several decades \cite{Nourbakhsh-1996-16261}. Shakey \cite{Nilsson84} (1966-1972) was the first robot that used Symbolic Planning. This robot was used for testing STRIPS \cite{Fikes71}, one of the first languages for AI Planning. Rhino robot approach \cite{Burgard88} generated an action plan to guide visitors through the Bonn museum, and Dervish \cite{Nourbakhsh_Powers_Birchfield_1995} automatically calculated his action plan at the AAAI 1994 Robot Competition, winning the Office Delivery event. However, the mechanism associated to these systems had computational limitations and the difficulty of integrating planning and action in robots working in complex and dynamic environments, which led researchers such as Arkin \cite{Arkin89} and Brooks \cite{Brooks91} to seek new techniques for generating robotic behaviors far from planning-based approaches.

 Planning Definition Domain Language (PDDL) \cite{pddl} was a significant step for the standardization of planning languages. This language, inspired by STRIPS, has become a standard that has driven this area of research. One of their releases, PDDL 2.1 \cite{PDD21}, was especially relevant as it included multiple enhancements, such as the ability to use temporal planning with durative actions, which introduces the concept of period of time into the PDDL framework.

Planning and Robotics are converging again through various planning systems and frameworks for generating behaviors for robots in recent years. ROSPlan \cite{ROSPlan19} is one of the most successful current planning frameworks, and is a reference in ROS \cite{ROS}. 
During each execution of a plan, ROSPlan dispatches the actions sequentially, one by one, activating a software component that carries them out. Each action in PDDL has a component that materializes the action in the robot. ROSPlan can be integrated into cognitive architectures \cite{Jonatan19}, providing planning capabilities. Our proposal is integrated into the ROS2 Planning System, which is the evolution of ROSPlan in ROS2. One of the improvements to ROSPlan is executing actions in parallel and different robots, thanks to a new activation protocol for the components that carry out a plan's actions. The approach proposed in this paper enables the control of this execution in parallel.

On the other hand, a Behavior Tree is a mathematical model of plan execution. They have probed more flexible and suitable than state machines for defining robots' simple behaviors. Behavior Trees come from the world of video games \cite{Marcotte2017BehaviorTF} and have become very popular in Robotics in recent years \cite{Marzinotto14}.  

In PlanSys2 these two tools have been combined into the software component that provides the service of implementing a PDDL action as a Behavior Tree. 

In \cite{Marzinotto14}, authors propose to solve an action plan by coding it into a behavior tree. Initially, they create a Behavior Tree with a single node containing the goal. Not being satisfied, at runtime, this node is expanded with the action whose effect is to achieve it. This action is a sequence composed of some requirements to be applied, with the action itself below. Recursively, for each requirement not satisfied, it is expanded including the action that satisfies it, along with its requirements. The authors show that with this strategy, the goal is reached after several expansions. Although this approach has proven feasible for simple plans, but it is not able to cope with complex plans, for instance, to deal with planning loops. 

SKIROS \cite{skiros2} is another behavior generation framework for ROS that uses planning. Its approach is different from ROSPlan. First, the knowledge base is based on ontologies written in OWL instead of in PDDL, from which the problem to be solved is generated. Second, it uses extended Behavior Trees to define the tasks that the robot can perform. These tasks contain abstract tasks that can be explored at run time by a scheduler. Although SKIROS publications describe a way to optimize the Behavior Tree to be executed, the available version only implements the sequential execution of tasks. 

Our approach starts from a PDDL plan, not from the expansion of skills as in SKIROS, so it will be possible to generate more complex plans. Our proposal is also capable of dealing with plans involving several robots, while SKIROS is a single-robot framework.

Zhou et al. \cite{zhou2019autonomous} present an engine for automatically generation of Behavior Trees (BT) form high level plans. Their approach transforms a plan into BT's with a continuous updated approach based on the task planning (and execution) algorithm {\em Hierarchical  task  and motion  Planning  in  the  Now}  (HPN)\cite{kaelbling2011hierarchical}. Our proposal goes beyond a domain-dependent scenario associated with hierarchical formalization. We generate a plausible BT, independent if those plans have a hierarchical decomposition or enhanced running some actions in parallel.

Besides, our proposal guarantees that all generated BT's have a solution, avoiding to introduce offline BT's checker as in Giunchiglia's  proposal \cite{giunchiglia2019conditional} for verifying the executability of their conditional BT's.

\section{Graph representation from plans}
\label{sec:graph}

PDDL defines a standard to code symbolic planning problems. The domain model defines what type of elements, and the predicates to describe them, for a given problem. It also specifies the requirements and the effects of the actions that can be applied in that environment. Using the rules that the model defines, problems can be solved by a calculated plan. A problem model is made up of a set of elements, predicates, and a goal to achieve through actions.

Let's consider the following domain, referenced as PDDL listing  \ref{pddl:domain1}. This domain, named \texttt{simple}, defines that objects can be of type \texttt{robot} or \texttt{room}). There are two different possible predicates (\texttt{robot\_at} and \texttt{connected}). When \texttt{grounded}, that is, when the parameters of the predicates are replaced for existing objects, predicates shape the knowledge base of the robot. The action \texttt{move}, once grounded in a plan, defines the requirements to be applied, and the effects of being executed. 


\begin{pddl}
\begin{verbatimtab}
(define (domain simple)
(:types robot room)
(:predicates
  (robot_at ?r - robot ?ro - room)
  (connected ?ro1 ?ro2 - room))
(:durative-action move
  :parameters (?r - robot ?r1 ?r2 - room)
  :duration ( = ?duration 5)
  :condition (and
    (at start(connected ?r1 ?r2))
    (at start(robot_at ?r ?r1)))
  :effect (and
    (at start(not(robot_at ?r ?r1)))
    (at end(robot_at ?r ?r2))))
)
\end{verbatimtab}
\caption{\label{pddl:domain1}The example domain.}
\end{pddl}

For example, PDDL listing \ref{pddl:problem1} defines four objects of the two permitted types by the model. The knowledge base also contains five predicates, grounded by objects, and complying with the constraints that the model imposes. The goal is the logic expression, composed by predicates, to be achieved by the plan. Any plan solver can calculate the plan to achieve, resulting the plan in PDDL listing \ref{pddl:plan1}. The first number is the time in which the action starts. 

\begin{pddl}
\begin{verbatimtab}
(define (problem problem_1)
(:domain simple)
(:objects
  r2d2 - robot
  bedroom living kitchen - room
)
(:init
  (robot_at r2d2 bedroom)
  (connected living bedroom)
  (connected bedroom living)
  (connected living kitchen)
  (connected kitchen living))
(:goal (and(robot_at r2d2 kitchen)))
)
\end{verbatimtab}
\caption{\label{pddl:problem1}The example problem.}
\end{pddl}

\begin{pddl}
\begin{verbatimtab}
0.00: (move r2d2 bedroom living)
5.00: (move r2d2 living kitchen)
\end{verbatimtab}
\caption{\label{pddl:plan1}The plan generated.}
\end{pddl}

 The generation of the plan is out of the scope of this work; instead, we focus on the PDDL domain and problem model that solvers use to generate plans. Using this information the \emph{planning graph} $G$ is built. It is a directed acyclic graph defined by the tuple $G=<A,C>$, where $A$ is the set of actions in a plan, corresponding to  nodes of the graph. $C$ is the set of directed arcs that represents the execution precedence of the actions. An \emph{action unit} $a_i \in A$ is a tuple $<t, action, R, E, >$. The first element, $t_{a_i}$, is the time in which the action starts, according to the plan. The second element, $action_{a_i}$, is the action itself to be executed. The set $R_{a_i}$ contains all the predicate that conform the requirements to execute the action, and $E_{a_i}$ is the set of predicates that will be added to the knowledge base by executing the action.

\begin{figure}[ht]
  \centering
  \includegraphics[width=0.7\linewidth]{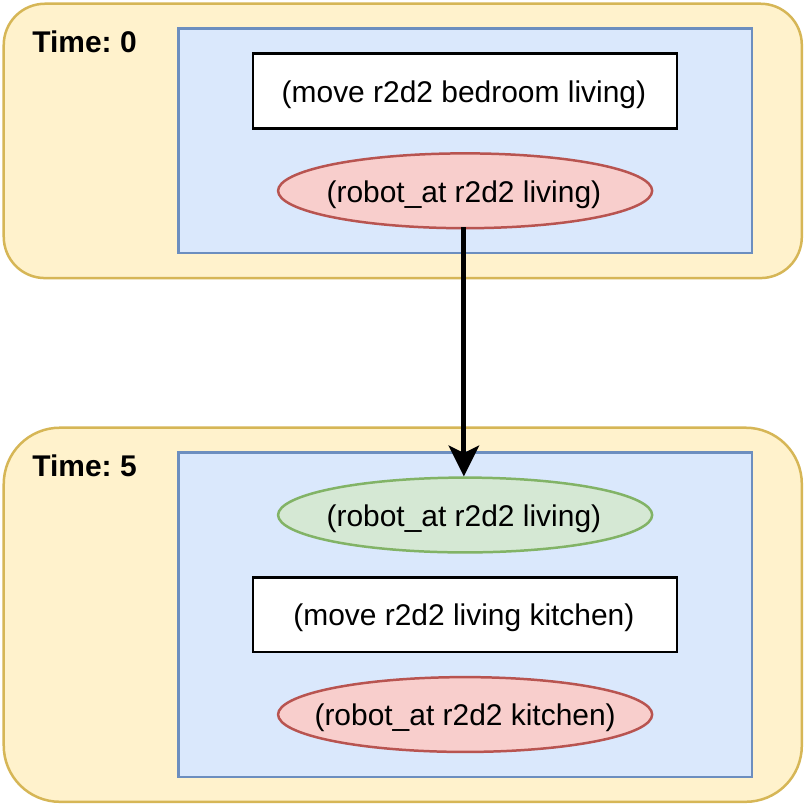}
  \caption{Planning graph corresponding to the plan in PDDL listing \ref{pddl:plan1}. Yellow boxes contain all the actions planned starting at the same time. Each action unit (blue boxes) contains requirements (green boxes) and effects (red boxes). Arcs in the graph link actions using the effect of the source action and the requirements of the target action.}
  \label{fig:graph1}
\end{figure}

The goal is obtaining a \emph{planning graph} as shown in Figure \ref{fig:graph1}, which corresponds to PDDL listing \ref{pddl:plan1}, from a plan in PDDL. This graph represents the different times included in the plan. The steps to build this graph are:
\begin{enumerate}
    \item Create an action unit $a_i$ for each action in the input plan. Initialize the action unit with the action $action_{a_i}$ to execute and the time $t_{a_i}$.
    \item For each action unit $a_i$, fill the requirements $R_{a_i}$ with the required grounded predicates to execute it. It is required to consult the model to obtain the predicates in the condition part of the action. Ground this predicate with the parameters of the action.
    \item Do the previous step with the effects $E_{a_i}$. Only positive effects (those that add a new predicate to the knowledge base) are considered. Most PDDL solver do not accept negative preconditions, so PDDL models should not include it to be portable. If used, the algorithm will work, but it can be inaccurate.
    \item Sort actions units according to time $t_{a_i}$. For each requirement $R_{a_i}$ in action $a_i$, find an effect $E_{a_j}$ of another action $a_j$ such as $R_{a_i} = E_{a_j}$, if $t_{a_i} > t_{a_j}$. When found, make an arc linking $R_{a_i} \to E_{a_j}$, implicitly linking $a_i \to a_j$.
    \item Identify the set of predicates $I$ in the initial problem that are not present in the effects of any action $a_i \in A$. For each action unit $a_i$, remove from $R_{a_i}$ the predicates in $I$, if any.
    \item Check that every $R_{a_i}, \forall a_i \in A$ is linked to any $E_{a_j}, \forall a_j \in A, i \ne j$.
\end{enumerate}

\begin{figure}[ht]
  \centering
  \includegraphics[width=0.5\linewidth]{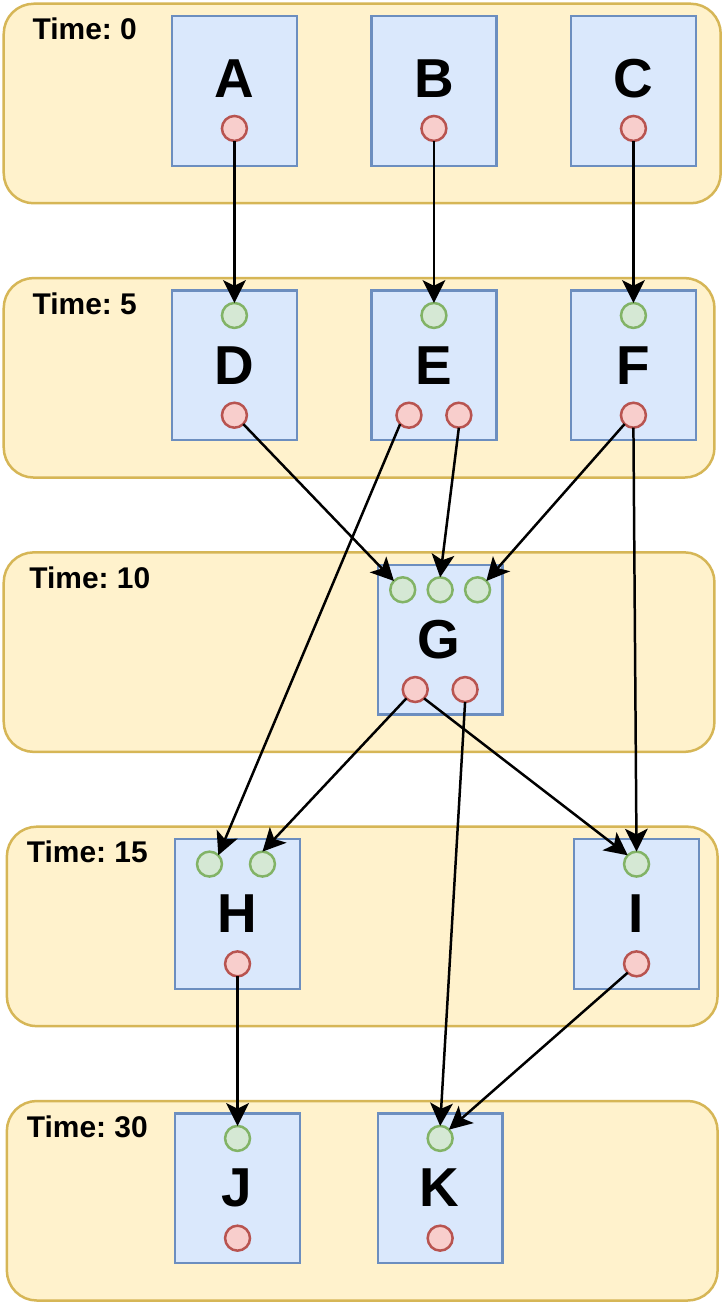}
  \caption{A complex planning graph.}
  \label{fig:graph2}
\end{figure}

Figure \ref{fig:graph2} shows a more complex plan. In this plan, there are actions that are planned to be executed in parallel. It can also be observed how an action can depend on the execution of other actions, or how an action allows the execution of several other actions.

We want to introduce the \emph{cardinality} term of an action unit $|a_i|$ as the number of arcs that go to, or start from $a_i$. The \emph{input cardinality} $|\to a_i|$ refers to the number of arcs received by $a_i$ from unique action units. Similarly, the \emph{output cardinality} $|a_i \to|$ indicates the number of arcs comming out from $a_i$ to unique action units. Several arcs from $a_i$ that connect to $a_j$ are considered the same connection, and $|a_i \to| = |\to a_j| = 1$.

\section{Behavior Tree building}
\label{sec:bt}

Behavior Trees are used to code the execution of actions through a tree that contains nodes, which are leaves if they are end nodes (without child nodes). The leaves represent actions to be carried out, or conditions to check. The rest of the nodes define the execution flow of the tree.

The main operation is {\em tick}. When a node is ticked, this operation can return three values:
\begin{description}
    \item[SUCCESS] Node function completed successfully.
    \item[FAILURE] Node function terminated, but failed.
    \item[RUNNING] The node function has not finished.
\end{description}

We tick the root of the Behavior Tree to execute it until it returns SUCCESS or FAILURE. This tick is propagated through the tree by the {\em control nodes}. There are several types of control nodes (sequence, fallback, decorators, mainly), but those relevant to this work are  three:
\begin{description}
    \item[Sequence] When a sequence node is ticked, it ticks its children. Each tick is made to a single child, starting with the one furthest to the left. When it returns SUCCESS, it is passed to the next child. If any of the children returns FAILURE, the sequence node returns FAILURE. The sequence node returns SUCCESS if all children have returned SUCCESS. In any other case, it returns RUNNING.
    \item[Parallel] This control node ticks all its children.  It returns SUCCESS if all its children return SUCCESS and FAILURE if any returns FAILURE. In any other case, it returns RUNNING. Using this node is the way actions are executed in parallel.
    \item[Condition] returns SUCCESS if the evaluation of the conditions results in true, and FAILURE for false.
\end{description}

\begin{figure}[ht]
  \centering
  \includegraphics[width=0.8\linewidth]{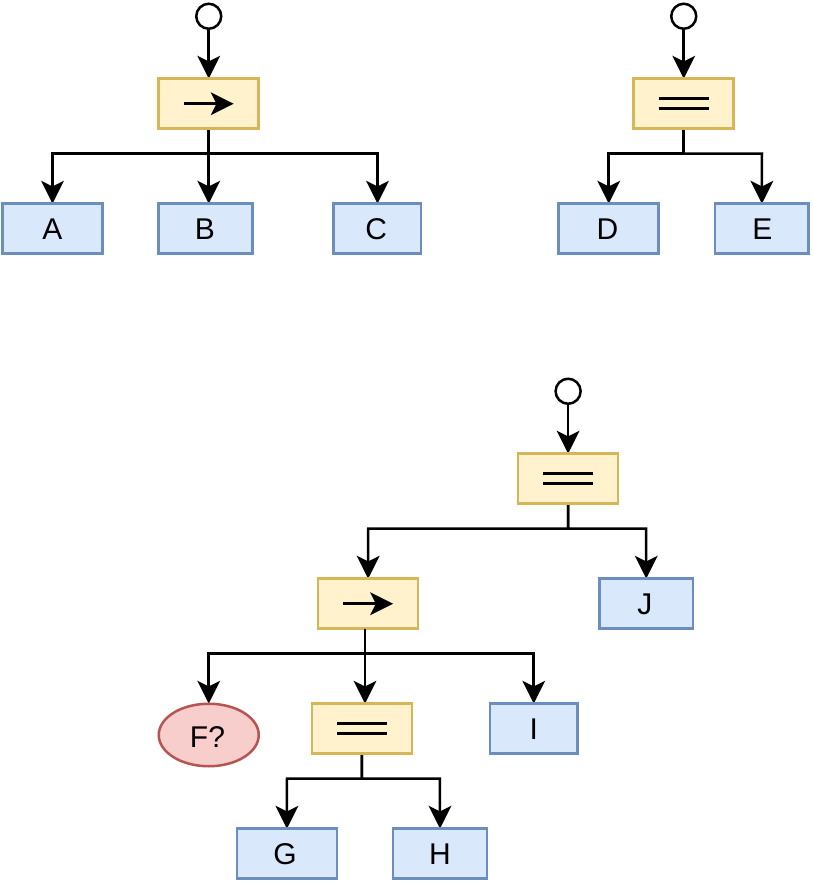}
  \caption{Three different behavior trees. Yellow rectangles are control nodes ($\to$ for sequences and $=$ for parallels). Red ellipse is a condition node, and blue rectangles are action nodes.}
  \label{fig:bt1}
\end{figure}

Figure \ref{fig:bt1} shows several examples of Behavior Trees. The Behavior Tree at the top left shows a tree with a sequence of actions. The tick to the root of the tree is propagated by ticking the sequence node. The execution of this node begins by ticking action $A$. If $A$ returns RUNNING, the action node returns RUNNING due to the tick at the root of the tree. When $A$ returns SUCCESS, the sequence node continues to return RUNNING since child nodes are left to run. The next tick to the sequence node will propagate to $B$. Once all the children have returned SUCCESS, the sequence node returns SUCCESS, making the tick at the root of the tree also SUCCESS, thus notifying that execution has finished. The Behavior Tree at the top right makes both actions $D$ and $E$ be ticked for each tick in the parallel node. The Behavior Tree at the bottom combines several sequence nodes, including a condition node, $F$, that can fail the execution of the sequence node, and so the entire tree.

\begin{figure}[ht]
  \centering
  \includegraphics[width=0.6\linewidth]{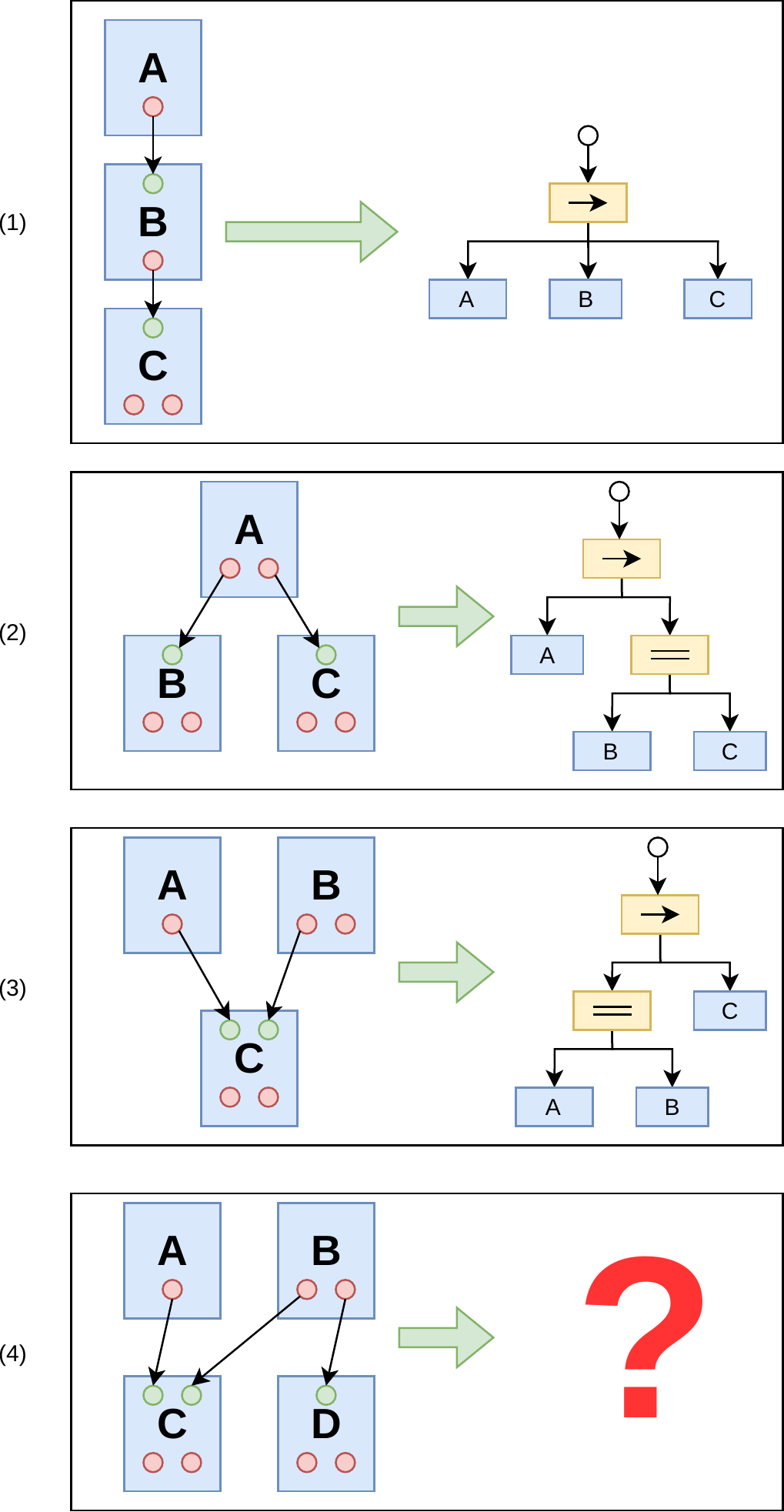}
  \caption{Translation from planning graph to Behavior Tree.}
  \label{fig:bt2}
\end{figure}

The planning graphs introduced in section \ref{sec:graph} can be transformed into Behavior Trees ready to be executed in most cases. Figure \ref{fig:bt2} presents several examples of these transformations:
\begin{enumerate}
    \item If an action unit $a_i$ whose $|a_i \to| = 1$ connects to another action unit $a_j$ whose $|\to a_j| = 1$ ($t_{a_i} < _{a_j}$ is guaranteed by graph construction), this is translated to a sequence structure.   
    \item If an action unit $a_i$, whose $|a_i \to| > 1$, diverges to more than one action units $a_{j,k,...}$, whose ($|\to a_{j,k,...}| = 1$), it transforms to a Behavior Tree with one sequence that executes $a_i$ before a parallel control node that contains action units $a_{j,k,...}$.
    \item If several action units $a_{j,k,...}$ with $|a_{j,k,...}\to| = 1$ converges to the same action unit $a_i$, it transforms to a Behavior Tree with one sequence that executes a parallel control node that contains action units $a_{j,k,...}$ before $a_i$.
    \item If an action unit $a_i$ with $|a_i\to| > 1$ connect with an action unit $a_j$ with $|a_j| > 1$, the transformation to Behavior Tree is not possible with no further information.
\end{enumerate}

\begin{figure}[ht]
  \centering
  \includegraphics[width=\linewidth]{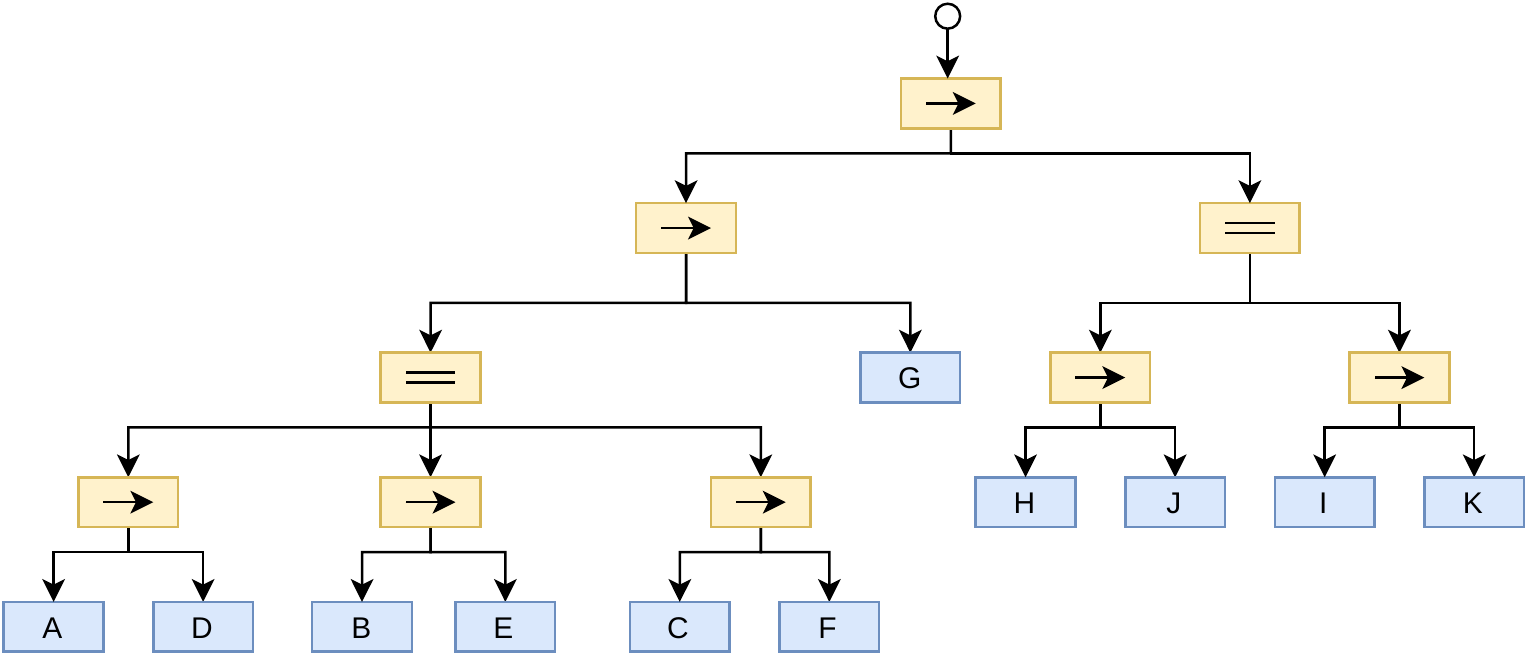}
  \caption{Translation into Behavior Tree of the planning graph shown in Figure \ref{fig:graph2}.}
  \label{fig:bt3}
\end{figure}

The last example in \ref{fig:bt2} presents the challenge faced. The planning graph shown in Figure \ref{fig:graph2} is translated into the tree of Figure \ref{fig:bt3},  including the problematic case just presented. Looking carefully to this planning graph, it  $E \to H$, $F \to I$ and $G \to K$ arcs can be ignored since other arcs really define the causal order of the actions: $E \to G \to H$, $F \to G \to I$ and $G \to I \to K$. If the arc $G \to I$, for example, would not exists, it would be hard to build the Behavior Tree.

\begin{figure*}[t]
  \centering
  \includegraphics[width=0.60\linewidth]{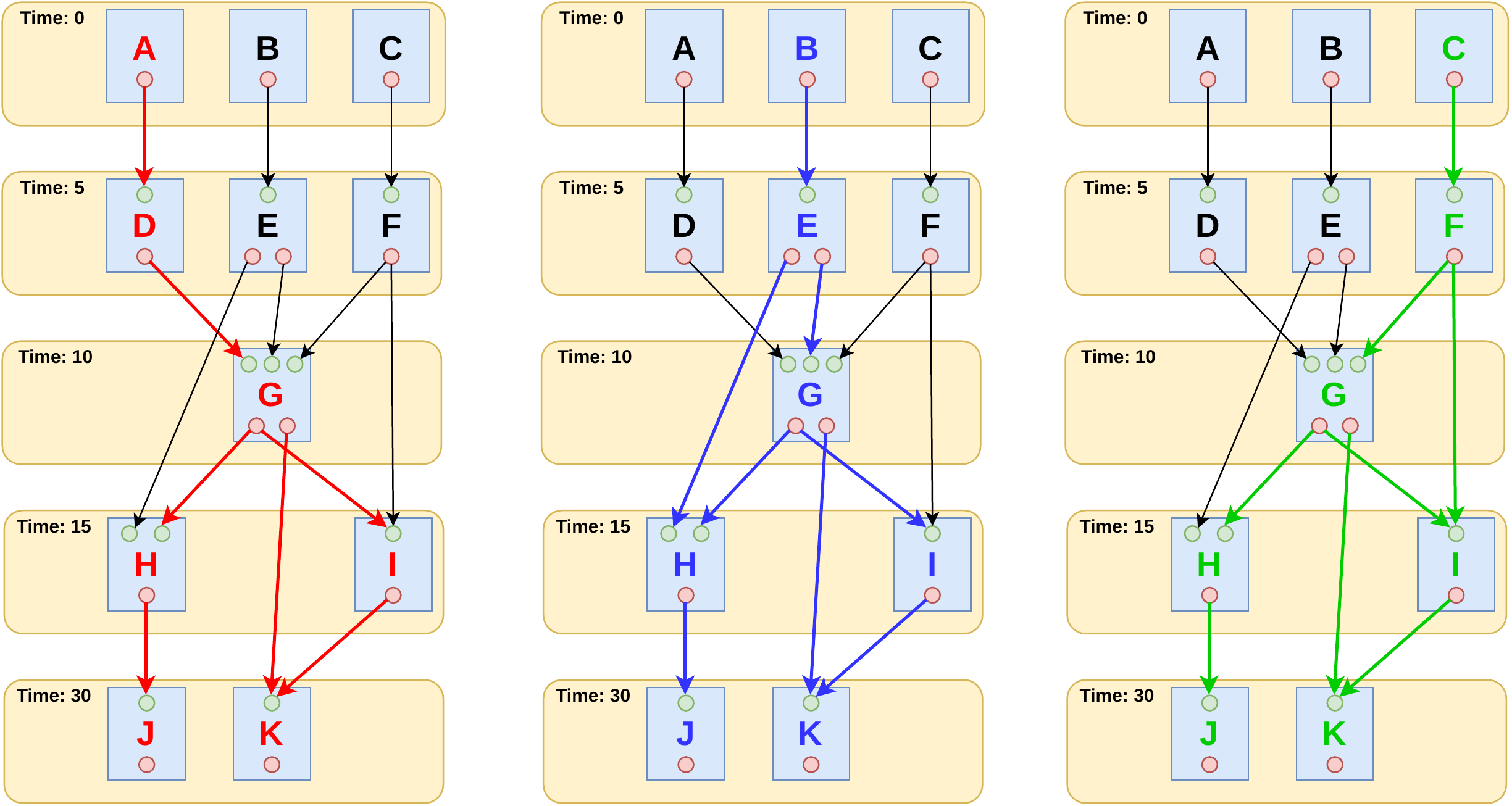}
  \caption{Execution flows of the planning graph shown in Figure \ref{fig:graph2}.}
  \label{fig:flow1}
\end{figure*}

The main contribution of this work is an algorithm that solves this problem. This algorithm  automatically builds a Behavior Tree from any type planning graph. It is based on three main concepts:

\begin{description}
    \item[Execution flows] An execution flow starts from an action $a_i$ node whose $R_{a_i} = \emptyset$, and add the actions in the path following the arc effect $\to$ requirement. In Figure \ref{fig:flow1} shows the three different flows in the planning graph shown in Figure \ref{fig:graph2}.
    \item[Singleton action] Each action unit $a_i$ in the generated Behavior Tree is a Singleton $S(a_i)$, i.e., it can appear in different points in the Behavior Tree, but it refers to the same action unit. If an action unit $a_i$ has already returned SUCCESS when executed in one branch, it will return SUCCESS if it is ticked in any other branch.
    \item[Waiting nodes] This control node refers to an action unit $a_i$, and denominates $W(a_i)$. It returns RUNNING if $a_i$ has never returned SUCCESS.
\end{description}

Using these three concepts, the Algorithm \ref{alg:as} to build a behavior tree can be summarized as:

\begin{description}
    \item [Lines 1] initializes the set $Flows$ with the action units that do not depend on the execution of any other action. Each one is the starting of an execution flow.
    \item [\texttt{Get\_Tree}] is a function that returns a subtree taking an action unit in the graph. The parameter $US$ is a set that accumulates the action already processed in the flow. 
    \item [Line 3] If there is more than one flow, the root has as child a Parallel control node, with one child per subtree starting in each flow. The second argument, $US$, is empty set in the first call of the algorithm.
    \item [Line 5] If there is only one flow, the root has as a child the subtree starting in the unique non-dependant action. $US$ is also empty set in this first call.
    \item [Line 8] Updates the set $US$ with the current action.
    \item [Line 9] Init the set $W$ with all the actions that has effects that $a$ requires, excluding those already processed in the flow.
    \item [Line 10] Initializes the set $Succ$ with all the actions that has requirements that $a$ satisfies.
    \item [Line 11-17] If the output cardinality of $a$ is 0; the leaf of the tree has been reached. It returns an Action node containing $a$. If $a$ requires the previous execution of other actions (line 12), it returns a Sequence Node. This sequence has as children the action preceded by so many Wait nodes as depending action.
    \item [Line 17-18] If the output cardinality of $a$ is 1; it means that execution is going through  a single sequence. The returning sequence contains:
        \begin{enumerate}
            \item For each action on which $a$ depends, a Wait node.
            \item The Action node containing $a$.
            \item A subtree with the child of $a$.
        \end{enumerate}
    \item [Line 17-18] If $a$ diverges in other parallels actions, the returning sequence is similar to the previous case, with the difference in the last element. Instead of a single subtree, a Parallel node is added, with as many children as actions that requires the execution of $a$ to execute.
\end{description}

\begin{algorithm}[h!]
\caption{Behavior Tree builder algorithm}\label{alg:as}
\begin{algorithmic}[1]

\State $Flows \gets a_i \in A,$ if $R_{a_i} = \emptyset$

\If{$|Flows| > 1$}
    \State tree $\gets$ Parallel(get\_tree($a_i$, $\emptyset$), $\forall a_i \in Flows$))
\Else
    \State tree $\gets$ get\_tree($a_i$, $\emptyset$), $a_i \in Flows$)
\EndIf

\Function{get\_tree}{$a$, $US$}
        \State $US \gets US \cup a$
        \State $W \gets a_j \in A,$ if $a_j \in R_{a}, a_j \notin US$
        \State $Succ \gets a_j \in A,$ if $a_j \ne a, a_j \in E_{a}$
        \If{$|a\to| == 0$}
            \If{$W \ne \emptyset$}
                \State return Sequence($Wait(a_j)$, Action($a$)), $\forall a_j \in W$
            \Else
                \State return Action($a$)
            \EndIf
        \ElsIf{$|a\to| == 1$}
            \State return Sequence$\left(
                \begin{array}{l}
                    Wait(a_j), \forall a_j \in W\\
                    Action(a), \\
                    get\_tree(a_j, US), \forall a_j \in Succ
                \end{array} \right)$ 
        \ElsIf{$|a_i\to| > 1$}
         \State return Sequence$\left(
                \begin{array}{l}
                    Wait(a_j), \forall a_j \in W\\
                    Action(a), \\
                    Parallel(get\_tree(a_j, US), \forall a_j \in Succ)
                \end{array} \right)$ 
        \EndIf
    \State 
\EndFunction

\end{algorithmic}
\end{algorithm}

\begin{figure}[t]
  \centering
  \includegraphics[width=0.7\linewidth]{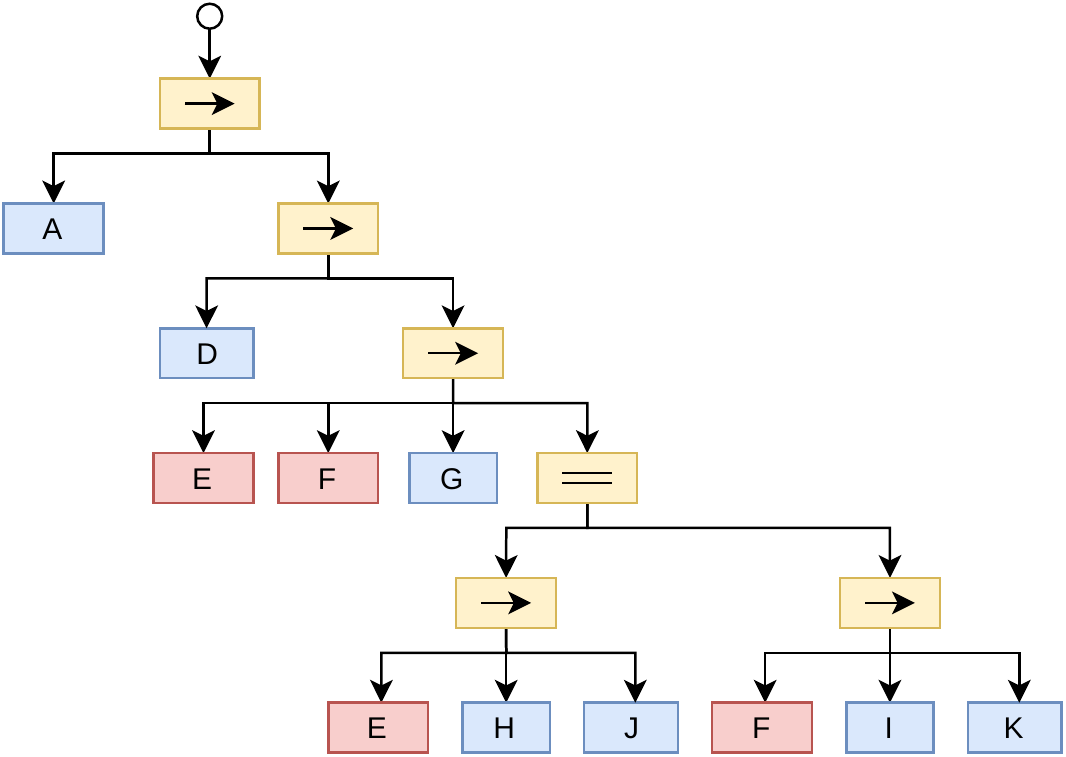}
  \caption{Behavior Tree generated from the red flow of Figure \ref{fig:flow1}. Red rectangles are Waiting Nodes.}
  \label{fig:bt4}
\end{figure}

Following Algorithm \ref{alg:as}, we can systematically create Behavior Trees from a planning Graph. Figure \ref{fig:bt4} shows the result of applying this algorithm to the red flow of Figure \ref{fig:flow1}. The other subtrees, flows blue and green, will also contain nodes $G, H, I, J$, and $K$, but each one is a Singleton (points to the same others subtree's homologous). A tick to $G$ in subtree from green flow, ticks the same action $G$ in subtree of the red flow. Waiting Nodes solve the situation in which an action $a_i$ in a execution flow requires a predicate that produces an action that not belongs to the same execution flow. 

\begin{figure}[ht]
  \centering
  \includegraphics[width=0.6\linewidth]{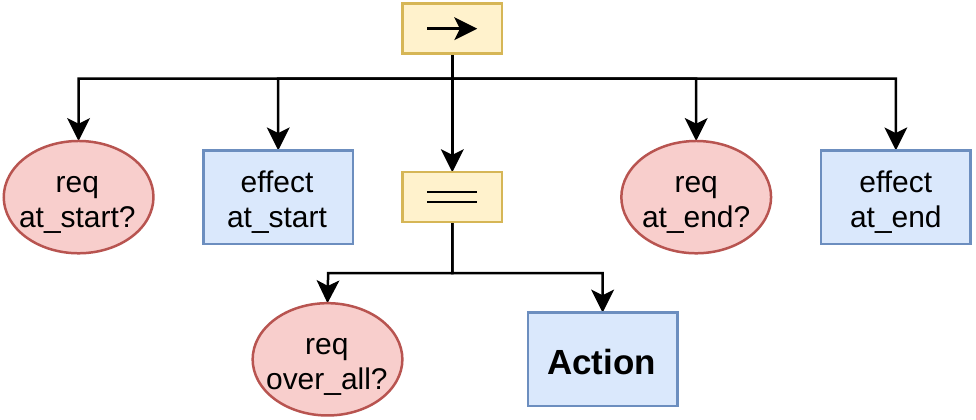}
  \caption{Expansion of each action unit.}
  \label{fig:bt5}
\end{figure}

Finally, each of the action nodes will be a subtree like the one shown in Figure \ref{fig:bt5}. In this subtree, the requirements checks will be made at runtime, as well as the application of the effects in their phase.

\section{Evaluation}
\label{sec:evaluation}

\begin{figure}[tb]
  \centering
  \includegraphics[width=\linewidth]{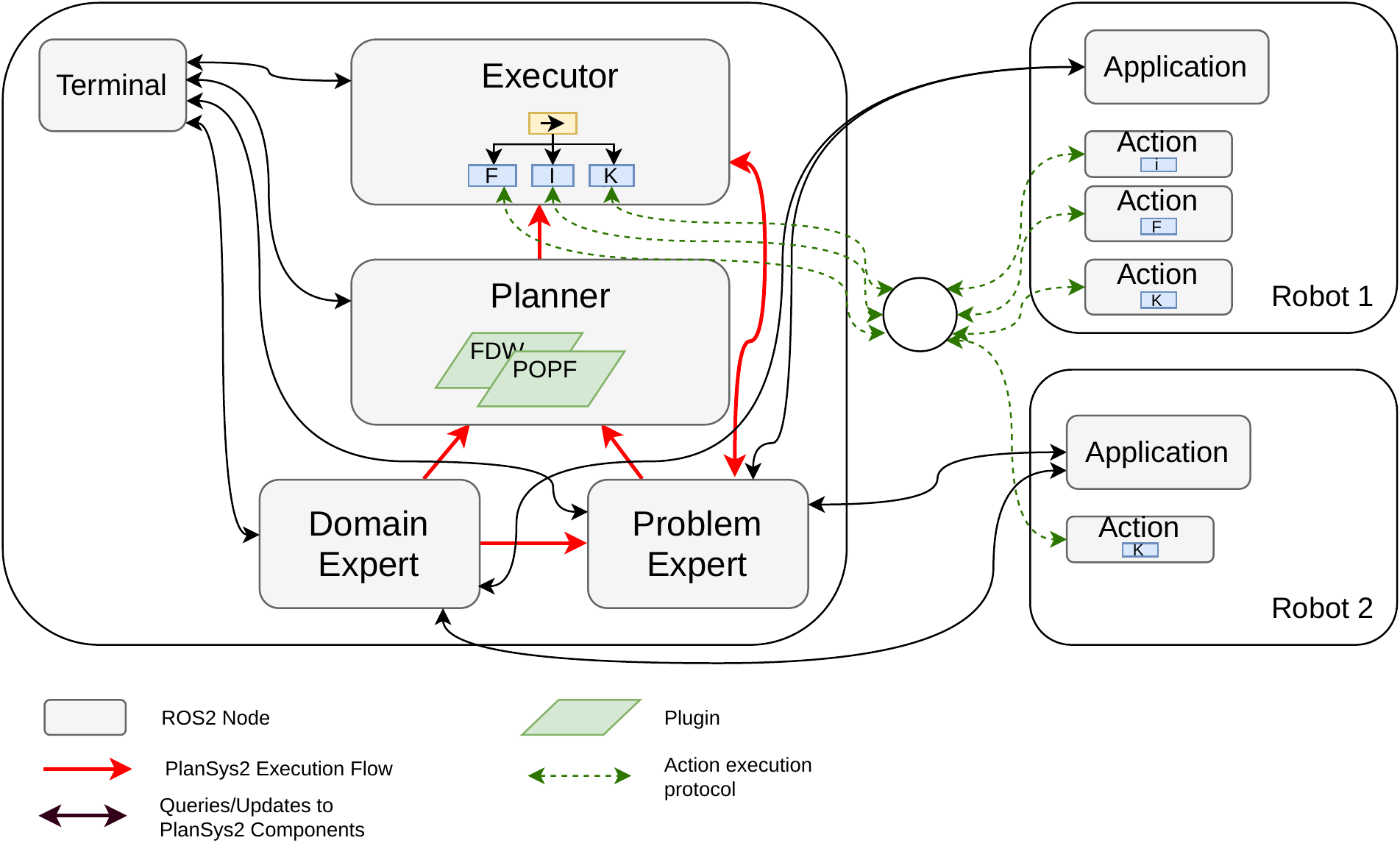}
  \caption{PlanSys2 architecture.}
  \label{fig:plansys2}
\end{figure}

The algorithm presented in this work has been implemented within PlanSys2. This planning system is made up of several modules, as shown in Figure \ref{fig:plansys2}. The Domain Expert is in charge of composing the PDDL domain that will be used in the plans. The Problem expert contains the knowledge base according to the PDDL domain. The Planner runs the PDDL plan solvers, which are available as plugins. POPF and Fast DownWard are currently available. The Executor takes a plan and is responsible for executing it. The Executor is where we build and run the Behavior Trees described in this work. The implementation of the actions is in different nodes, even in other robots. There is a Hub where both requests for the execution of actions and feedback on their execution concur. The Action nodes' execution in the Behavior Tree uses this hub, when ticked, to request an action's execution and obtain its execution status. Applications access the knowledge base to update and consult its content.

PlanSys2 contains multiple improvements over ROSPlan:
\begin{itemize}
    \item Modular design. This lets developers try different approaches for each of the modules.
    \item Composable model. The PDDL model can be composed of several independent models of different subsystems. Each subsystem contributes with its piece of PDDL.
    \item Multiple instances. Multiple instances of PlanSys2 can be run, each in a different ROS namespace. This is required for multi-robot support.
    \item Protocol for the execution of actions based on bidding which allows:
    \begin{itemize}
        \item The execution of more than one action simultaneously by the nodes that implement the actions.
        \item Conditional execution of actions depending on the arguments. For example, two robots R2D2 and C3Po, both implement the move action, similar to the one shown in PDDL listing 1. The node that implements move in R2D2 would only agree to execute the action if the first argument is R2D2.
    \end{itemize}
    \item New tools like \texttt{Terminal}, which provides a shell for introspection and control of the planning system.
    \item Action nodes that can be coded as Behavior Trees for execution.
\end{itemize}

\begin{figure}[ht]
  \centering
  \includegraphics[width=\linewidth]{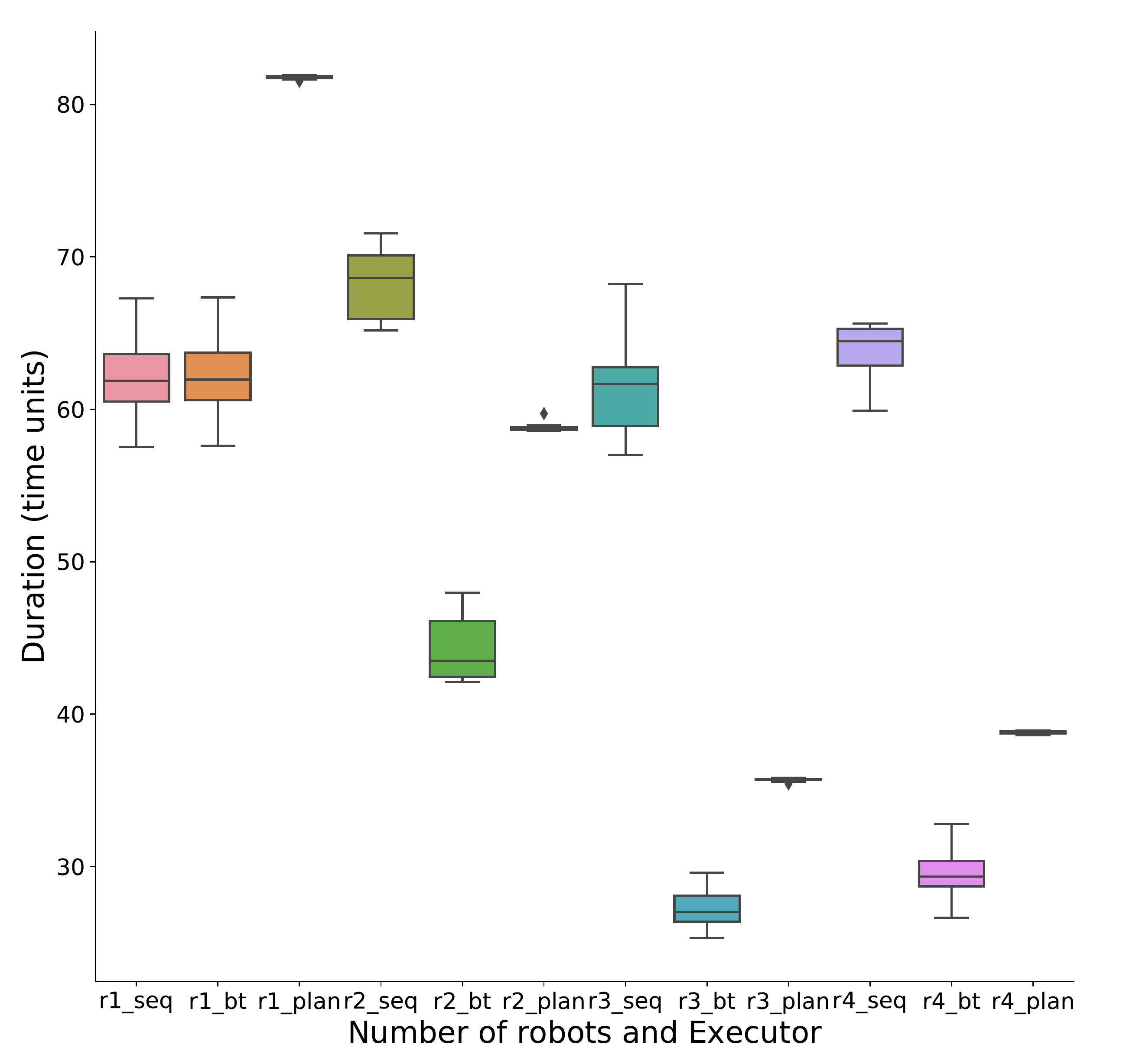}
  \caption{Starting from left, total time for execution with one robot (r1\_seq, r1\_bt, r1\_plan), two robots (r2\_seq, r2\_bt, r2\_plan), three robots (r3\_seq, r3\_bt, r3\_plan) and four robots (r4\_seq, r4\_bt, r4\_plan).}
  \label{fig:sciroc}
\end{figure}

PlanSys2 has been design with competitions as SciROC\footnote{https://sciroc.org/} in mind. In this competition, robots have to face various social tests, such as being a waiter in a restaurant (Figure \ref{fig:sciroc}) or going up in an elevator with more people. In order to evaluate the approach proposed in this work, it has been compared with two other execution models in a Sciroc scenario using as metric the  time required to carry out a plan. This two execution models are:

\begin{description}
    \item[Planner] This directly uses the output of the planner, executing the actions in the time stamp specified in the plan. These times are calculated from the duration specified in each action definition. Using this time is realistic because, in critical real-time systems, the scheduling uses the maximum expected time, and the actions are launched in these times. Normally, actions take less time to execute, but this execution model does not take advantage of this.
    \item[Sequential] The planner specifies an order, and the actions are sequentially executed in this order. A priori, this is the worst option, but it has been included in the comparison because it is currently the strategy in most of the systems that currently use planning for robots.
\end{description}

The scenario used  in this evaluation is the SciRoc Restaurant test, but including one or more robots to see how execution can be optimized by performing various actions in parallel. In this scenario, three tables must be serviced. The robot attends at a table when:
\begin{enumerate}
    \item The robot is asked for an order.
    \item The robot serves the food on the table. 
    \item The people on the table has finished eating.
    \item The robot is receiving the payment.
\end{enumerate}

The PDDL listing \ref{pddl:sciroc_plan} shows the plan generated for three robots collaborating in this task. On the left side of this plan, it is indicated when the action should start. On the right side is shown the duration of each action.

\begin{pddl}
\begin{verbatimtab}
0	(move robot1 kitchen table_a)	2
0	(move robot2 kitchen table_b)	2
0	(move robot3 kitchen table_c)	2
2	(ask_order robot2 table_b)	3
2	(ask_order robot3 table_c)	3
2	(ask_order robot1 table_a)	3
5	(move robot1 table_a kitchen)	2
5	(move robot2 table_b kitchen)	2
5	(move robot3 table_c kitchen)	2
7	(prepare_order robot1 kitchen table_a)	5
12	(move robot1 kitchen table_a)	2
12	(prepare_order robot2 kitchen table_b)	5
14	(serve robot1 table_a)	1
15	(wait_table table_a)	10
17	(move robot2 kitchen table_b)	2
17	(prepare_order robot3 kitchen table_c)	5
19	(serve robot2 table_b)	1
20	(wait_table table_b)	10
22	(move robot3 kitchen table_c)	2
24	(serve robot3 table_c)	1
25	(collect_payment robot1 table_a)	1
25	(wait_table table_c)	10
26	(move robot1 table_a table_b)	2
28	(move robot1 table_b table_c)	2
30	(collect_payment robot2 table_b)	1
35	(collect_payment robot1 table_c)	1
\end{verbatimtab}
\caption{\label{pddl:sciroc_plan}The plan generated to attend the tables with three robots.}
\end{pddl}

Each plan generated by the solver has been executed ten times by each of the three executors mentioned. In each iteration, each action's duration time has been generated randomly from a normal distribution $\mathcal{N}(\frac{3}{4}t, \frac{1}{8}t$), where $t$ is the maximum expected duration of the action. This approach is realistic since, in critical systems, an action's duration is modeled, adjusting it to the maximum expected duration. This assumption can be considered  real and appropriate for our evaluation. The results shown below compare the mean of the ten iterations per experiment.

\begin{figure}[ht]
    \centering
    \includegraphics[width=0.5\textwidth]{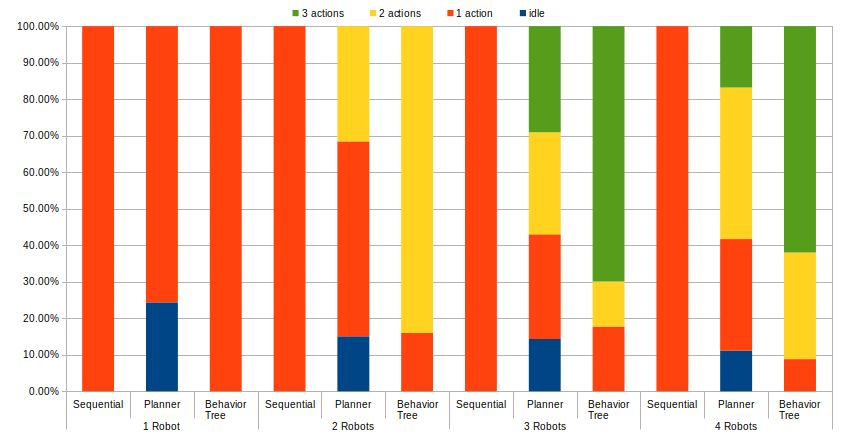}
  \caption{Percentage of time Idle, executing 1 action, 2 actions and 3 actions in the SciRoc scenario.}
  \label{fig:experimentaldata}
\end{figure}

	

Figure \ref{fig:experimentaldata} shows the percentage of time that the executor is running one, two or three actions, or is idle. The ratio in which the system is idle vs. running is significant in  execution time, always taking into accout what the planner indicated. An action can end before its maximum expected duration, and actions do not start until the plan's time point. In the case of a single robot, in the execution according to the plan, the system is idle around 24.32\%. The plan's sequential execution shows that there is always an action running, regardless of the number of robots. This is what actually happens in most current planning systems, which carry out actions following this strategy. The proposal described in this paper maximizes the parallel execution of actions. The highest percentage is got by the configuration that takes advantage of the maximum number of robots available.

This optimization in the execution of actions is reflected in the complete plan's execution time, shown in Figure \ref{fig:sciroc}.  As more robots are added, sequential execution maintains the same duration, since it does not optimize anything. Execution time according to plan decreases when more robots are included. The execution based on Behavior Tree shows the best performance in all configurations. The execution based on the planner is the most deterministic one due to planning that takes into account the maximum times per action. It should be noted that when resources are increased, in this case, robots, but there are no tasks (tables in this scenario) for every robot, performance deteriorates. Too many robots impede in our experiment, as seen in Figure \ref{fig:experimentaldata} and rightmost part of Figure \ref{fig:sciroc}.

\section{Conclusions}
\label{sec:conclusions}

This paper presents a proposal for using Behavior Trees to execute plans generated by a PDDL-based AI planner. Coding a plan as Behavior Tree is a compact way to represent and execute a robot action plan. Major contribution of the paper is the algorithm capable of transforming any plan into a Behavior Tree in a systematic way. This solution creates a planning graph  from the plan and makes the tree recursively. Different types  of nodes are used to build  the Behavior Tree such as the singleton action node and the wait node to improve the efficiency of parallel excution of actions. The generated Behavior Tree is so optimized to execute in parallel all the possible actions in a plan, preserving the causal relationships of the actions. Another contribution is the execution an action as soon as its requirements are available, even before established in the plan.


This algorithm and the Behavior Trees executor have been included in the plan execution module of the ROS2 Planning System. Besides, it has the ability to execute plans on multiple robots collaboratively showing a positive impact in a real competition test with a multirobot variant.

Using Behavior Trees opens the door for future research. In particular,  the monitorization of the execution of plans with tools such as Groot could be implemented. New types of nodes with particular functionalities could also be included in the Behavior Trees. These nodes can control aspects such as safety or quality of service aspects.




\bibliographystyle{ACM-Reference-Format} 
\bibliography{sample}


\end{document}